\documentclass[runningheads]{llncs}
\usepackage{graphicx}
\usepackage{multirow}
\usepackage{subfigure}
\begin{document}
\title{Towards Dialogue-based Navigation with  Multivariate Adaptation driven by Intention and Politeness for Social Robots}
\titlerunning{Towards Dialogue-based Navigation for Social Robots based on Politeness}

\author{Chandrakant Bothe\inst{1}\and
Fernando Garcia\inst{2}\and
Arturo Cruz Maya\inst{2}\and \\
Amit Kumar Pandey\inst{2}\and
Stefan Wermter\inst{1}
}

\authorrunning{C. Bothe et al.}
\institute{Knowledge Technology, Department of Informatics\\
University of Hamburg, Hamburg, Germany\\
\email{\{bothe, wermter\}@informatik.uni-hamburg.de} \and
SoftBank Robotics Europe, Paris, France \\
\email{\{ferran.garcia, arturo.cruzmaya, akpandey\}@softbankrobotics.com} 
}
\maketitle  % typeset the header of the contribution
\begin{abstract}
Service robots need to show appropriate social behaviour in order to be deployed in social environments such as healthcare, education, retail, etc.  
Some of the main capabilities that robots should have are navigation and conversational skills. 
If the person is impatient, the person might want a robot to navigate faster and vice versa. 
Linguistic features that indicate politeness can provide social cues about a person's patient and impatient behaviour.
The novelty presented in this paper is to dynamically incorporate politeness in robotic dialogue systems for navigation.
Understanding the politeness in users' speech can be used to modulate the robot behaviour and responses. 
Therefore, we developed a dialogue system to navigate in an indoor environment, which produces different robot behaviours and responses based on users' intention and degree of politeness.
We deploy and test our system with the Pepper robot that adapts to the changes in user's politeness.
%the results where the robot adapts to the changes in user's politeness during the interaction.
%with two scenarios when a user is polite and impolite

%ADD: Pepper robot name in abstract...

%The abstract should briefly summarize the contents of the paper in 15--250 words.

\end{abstract}

\section{Introduction}
%In this work, we address the problem of developing a dialogue system with multivariate behavioral adaptation based on sociolinguistic aspects such as politeness
%as a preliminary step towards potentially learning safety concepts 
%for safer 
%for human-robot interaction (HRI).
%Politeness affects the perception of the users towards their patience during the interaction.
The perception of politeness of a user can be a reflection of their patience during interaction.
In addition to other factors such as the robot appearance, robot behaviour is a crucial aspect for their acceptance.
%This work is a step towards developing a dialogue system for safe and pro-active HRI.
%The sociolinguistic factor like politeness plays an important role in whether social interaction goes well or poorly.
Hence, politeness cues are intimately related to the dynamics of behavior and interaction \cite{levinson1987politeness,P13dan2013compPoliteness_acl,holmes2015power,srinivasan2016help}.
It is useful for adapting to the dynamic tension \cite{rogers2003reconceptualizing} that occurs as a user tries to maintain a sufficient degree of politeness while interacting with the robot. 
For example, sentence-initial \textit{you} or an action directive verb can be impolite \textit{``You need to show..."} or \textit{``Show me the..."}, whereas sentence-medial \textit{you} or sentence-initial \textit{could} or \textit{would} often indicates the politeness like in these sentences \textit{``Could you show me..."} or \textit{``Would you take me to..."}.

%how conventional politeness dimensions, such as deference, solidarity, and non-imposition
Multivariate adaptive and affective dialogue systems based on linguistic features have been subject to previous research \cite{adam2016social,fong2003survey,shi2018sentiment}. 
The effect of politeness on the conversation is prominent and it has been researched in the sociolinguist community \cite{P13dan2013compPoliteness_acl,holmes2015power,rogers2003reconceptualizing}.
The effect of such a feature on human-robot interaction (HRI) has been a subject of study with various aspects: an impolite vs. a polite robot playing a game \cite{castro2016effects}, in determining social robot acceptance with multi-cultural background people \cite{salem2014marhaba}, making robots sociable and to achieve safe HRI \cite{fong2003survey}.
Hence, a robot that can recognize the intention of the user during interaction should also adapt to the human's linguistic behavioural changes.
For example, different sociolinguistic features such as politeness, emotion, sentiment, etc. represent the user's behavioural dynamics. 
If the user's utterance is impolite, then he/she might be in a hurry and vice versa.
In such cases, the robot might need to change its behaviour or even alter some actions or speed up or down the movements.
We develop a modular dialogue system (DS) that can process such features and make a robot to adapt accordingly. 
The natural language understanding part of the DS uses recurrent neural networks (RNNs) \cite{ultes2017pydial,yang2017end_icassp} and the Snips library \cite{coucke2018snips} for extracting the structured information from the user input.
The politeness detection is learned from the human-annotated corpus and fine-tuned for the scenario-specific data.
The navigation of the Pepper robot is achieved by using the NAOqi framework.
The robot behaviour and responses are driven by dialogue flow module using the intention and politeness.
The main contributions of the present work towards bridging the gap between sociolinguistic research and HRI community are:

- developing a dialogue-based navigation for incorporating politeness, and

- incorporating sociolinguistic features for robotic behavioural modelling.

%\section{Related Work}

%Multivariate adaptive dialogue system based on linguistic features have been subject to research for a long time, a few to mention \cite{adam2016social,shi2018sentiment,fong2003survey}. 
%The effect of politeness on the conversation is prominent and it is been researched by sociolinguist research community \cite{rogers2003reconceptualizing,P13dan2013compPoliteness_acl,holmes2015power}.
%The effect of politeness on the human-robot interaction has been in research with different aspects, an impolite vs. a polite robot playing game \cite{castro2016effects}, in determining social robot acceptance with multi-cultural background people \cite{salem2014marhaba}, making robots sociable and to achieve safe human-robot interaction \cite{fong2003survey}.

To the best of our knowledge, our system is the first dialogue-based navigation that incorporates politeness as an important social cue to drive the robot behaviour and responses. 

\begin{figure}[t]
\includegraphics[width=10cm]{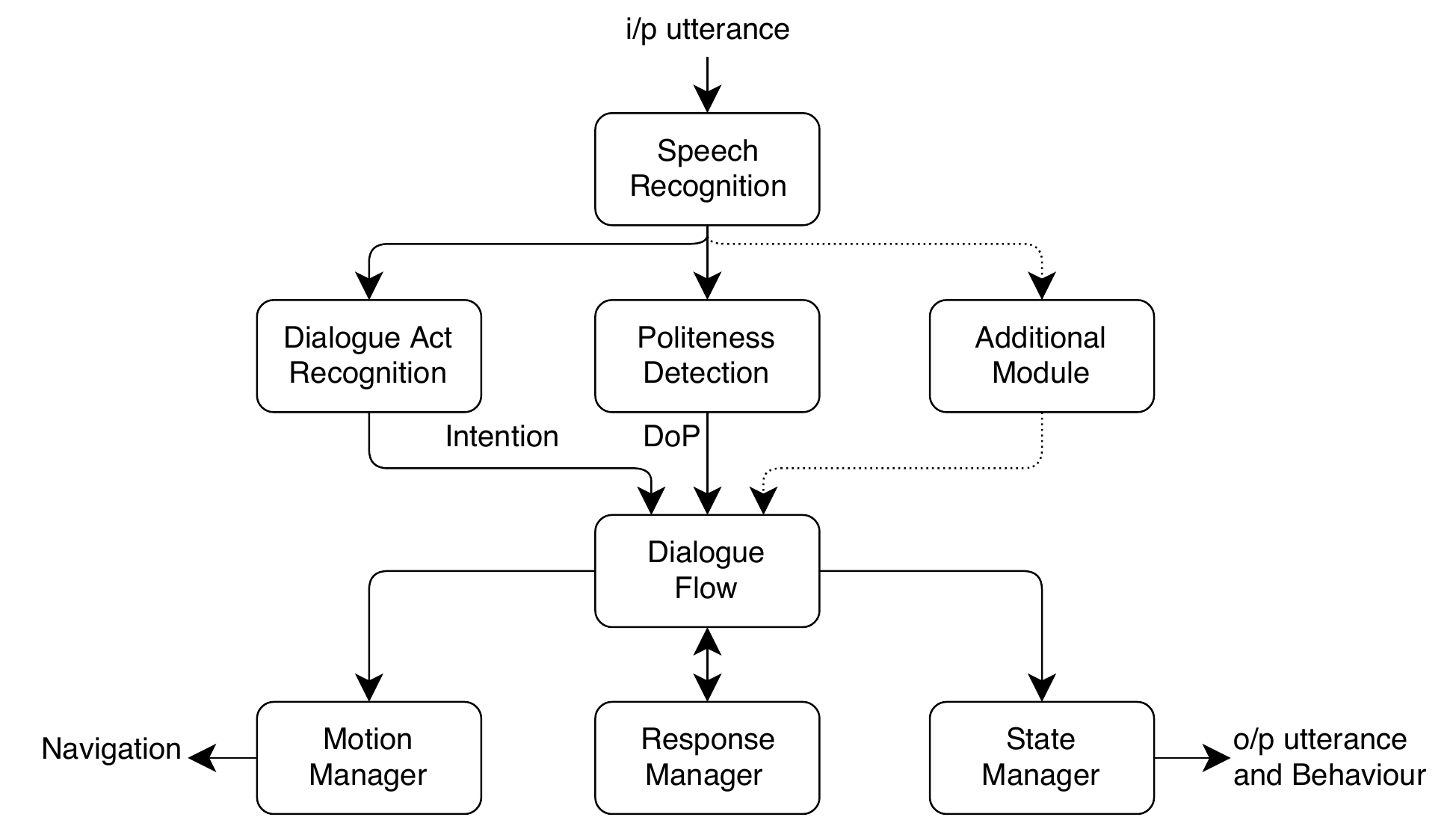}
\centering
\caption{The overall architecture of the dialogue system. DoP: degree of politeness.} 
\label{fig:overall_arch}
%\vspace{-20pt}
\end{figure}

\section{Approach}

We propose a dialogue system which takes into account the degree of politeness as a factor that affects the conversation flow and the robot behaviour.
The dialogue system takes the intention and other information, for example, slot-value pairs (see details in Section \ref{section:da_recg}) into account to understand the input utterance \cite{ultes2017pydial}.
% * <crbothe255@gmail.com> 2018-08-23T15:21:18.027Z:
% 
% add scalability : we use Politeness module but scalable to other extend
% 
% ^.
However, in the proposed model, the dialogue system processes sociolinguistic features to make inferences regarding the input utterance.
The overall architecture is shown in Figure \ref{fig:overall_arch}.
The proposed system is customizable to any extent as the robot is controlled using a client-server architecture.
The state and motion managers are wrapped into an application programming interface (API) as a server \cite{grinberg2018flask} and communicated via the dialogue flow module.
The dialogue system can be accessed also if the robot is not connected to the server. 
%s, for example, one could input and output in the form of text.

\section{Dialogue System}

%As it is mentioned before the dialogue system (DS) is modular and can also be used independently of a robot.
%The main part of the DS is the language understanding module which consists of dialogue act recognition, politeness detection, and additional modules.
%The central part of the DS is the dialogue flow module which takes care of the flow of utterances and sending commands to the robot.
%The response manager is responsible to produce an appropriate response given the intention and the degree of politeness.

\subsection{Natural Language Understanding}

The input speech from a user is converted into text using the speech recognition module (from NAOqi).
The language understanding module takes the converted input utterance forms a symbolic representation and provides the degree of politeness.
The dialogue act recognition module is used to extract the symbolic representation and politeness detection module to detect the degree of politeness of that input utterance. 
%we do not restrict the language understanding to dialogue act recognition only but extend it to politeness detection which can provide the degree of politeness given the input utterance.

\subsubsection{Dialogue Act Recognition Module}
\label{section:da_recg}
The dialogue act (DA) recognition is a crucial process in the dialogue system.
The task is to decode the input utterance and form the symbolic representation, such as dialogue acts and slot-value pairs.
For example, the utterance \textit{``could you please show me the retail department"} can be decoded as \textit{\{da : TakeToPlace, room : retail\} } where \textit{da} represents the dialogue act or intention, \textit{room} is a slot and \textit{retail} its value.
We created a dataset for the given scenario to be able to drive the conversation (some examples are given in Table \ref{table:examples}). 
The following methods are used in conjunction for robustness by validating one another based on heuristics of their confidence values:
%The methods to extract structured information from the input utterance: 

\begin{figure}[t]
\includegraphics[width=11cm]{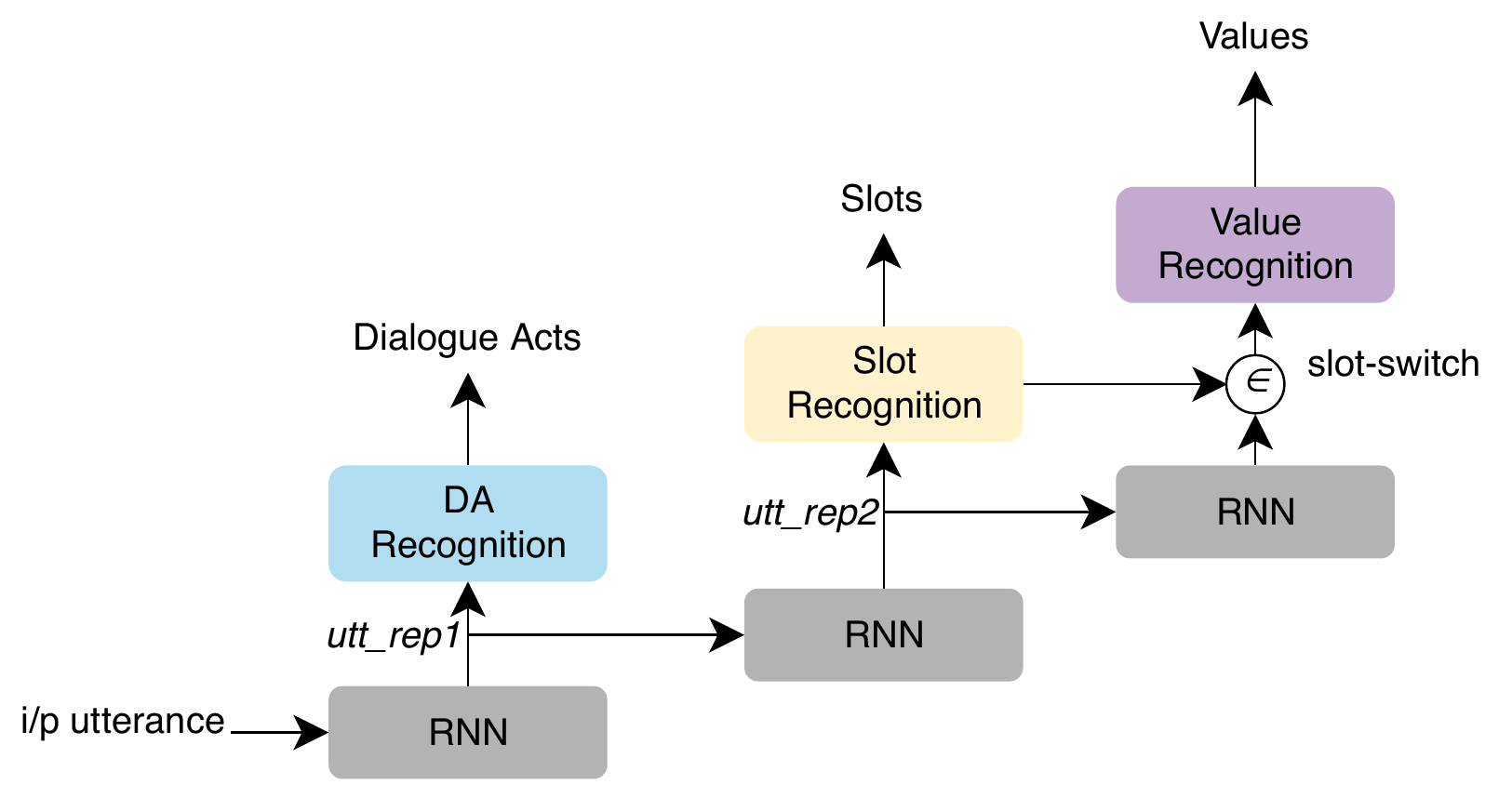}
\centering
\caption{Dialogue acts and slot-value pairs recognition using RNNs.} 
\label{fig:slu_model}
\end{figure}

\textbf{(1) Dialogue act recognition using RNNs: } 
The architecture is shown in Figure \ref{fig:slu_model}, where RNNs are used in a hierarchical fashion to learn the dialogue acts and slot-value pairs \cite{bothe2018discourse,kumar2017dasl,yang2017end_icassp}.
The RNNs are better at encoding the contextual and sequential information in the utterance \cite{kumar2017dasl}.
The dialogue acts are classified with the first layer of the RNN, preserving the utterance representation \textit{utt\_rep1}.
\textit{utt\_rep1} is then used to recognize slots on the next layer of the RNN, producing a new utterance representation \textit{utt\_rep2}.
The values of slots are learned with next layer of the RNN using \textit{utt\_rep2} and the detected slot as a switch ($\in$) for the belonging values learned in this layer (see the output in Figure \ref{fig:slu_example} for better understanding).
We fit the model to the data and use the trained model for inference \cite{bothe2018discourse}.

\textbf{(2) Snips Natural Language Understanding (NLU) Engine: } Snips NLU Engine\footnote{\url{https://snips-nlu.readthedocs.io}} is an open source Python library that uses two approaches: (a) a deterministic parser and (b) a probabilistic parser \cite{coucke2018snips}. 
The deterministic parser is basically a pattern matching mechanism which uses regular expressions to parse the input utterance.
The probabilistic parser uses logistic regression for intent classification and conditional random fields (CRFs) for slot filling.
For the given input utterance, the engine provides the intention and slot-value pairs.
%The library is open source and it is available with the documentation.

\subsubsection{Politeness Detection Module}

The politeness detection 
%is one of the key processes in the DS architecture as it drives the response and robot behaviour.
%This 
module takes the input utterance as an input and computes its degree of politeness.
An RNN is used to learn the degree of politeness from Stanford Politeness Corpus\footnote{\url{https://www.cs.cornell.edu/\~ cristian/Politeness.html}}\cite{P13dan2013compPoliteness_acl}.
We fine-tune the trained model for the dataset (mentioned in the previous section) that is created for the particular scenario to minimize uncertainty in prediction.
The degree of politeness (DoP) varies from 1 to -1 (very polite to very impolite).
For the sake of conceptual and computational simplicity, we discretized them into categories: polite (1), neutral (0) and impolite (-1); see the examples below:

\begin{verbatim}
DoP   Class     Utterance
 1    polite    Could you please show me the education department?
 0    neutral   Can you show me the education department?
-1    impolite  Show me the education department.
\end{verbatim}

% \usepackage{multirow}
\begin{table}[t]
\caption{Examples of dialogue act and slot-value pairs}
\label{table:examples}
\begin{tabular}{|c|l|c|c|}
\hline
\textbf{Dialogue acts} & \multicolumn{1}{c|}{\textbf{Examples}}                                                                                                                                 & \textbf{Slots}             & \textbf{Values}                                                                            \\ \hline
Greeting               & \textit{\begin{tabular}[c]{@{}l@{}}Hello.\\ Hi, how are you?\end{tabular}}                                                                                             & no\_slot                   & no\_value                                                                                  \\ \hline
Thanking               & \textit{\begin{tabular}[c]{@{}l@{}}Thank you.\\ Thank you very much.\end{tabular}}                                                                                     & no\_slot                   & no\_value                                                                                  \\ \hline
TakeToPlace            & \textit{\begin{tabular}[c]{@{}l@{}}Could you show me the education department?\\ Take me to the retail section.\\ Can you take me to tourism department?\end{tabular}} & room                       & \begin{tabular}[c]{@{}c@{}}retail\\ education\\ tourism\end{tabular}         \\ \hline
MoveRobot              & \textit{\begin{tabular}[c]{@{}l@{}}Please go ahead.\\ Could you move ahead?\\Go back please.\end{tabular}}                                                                              & \multirow{2}{*}{direction} & \multirow{2}{*}{\begin{tabular}[c]{@{}c@{}}forward\\backward\\right\\left\end{tabular}} \\ \cline{1-2}
TurnRobot              & \textit{\begin{tabular}[c]{@{}l@{}}Can you turn right?\\ Could you turn left.\\ \end{tabular}}            &             &                   \\ \hline
Accept                 & \textit{\begin{tabular}[c]{@{}l@{}}Yes, I would like to visit.\end{tabular}}                                                                                    & no\_slot                   & no\_value                                                                                  \\ \hline
AbortRobot             & \textit{stop, wait, be careful...}                                                                                                                                     & no\_slot                   & no\_value                                                                                  \\ \hline
\end{tabular}
\end{table}

\subsubsection{Additional Module}
This module is open to adding additional sociolinguistic features such as sentiment, emotion, etc. 
Adding more features can increase the complexity of the dialogue system. 
However, it could be useful in some cases to incorporate multiple features and modalities to produce the required behaviour.

\subsection{Dialogue Flow}
\label{section:dialogue_flow}

The dialogue flow is a central engine of the system which communicates with most of the modules.
It is implemented as a main function to drive the DS.
A rule-based and probabilistic belief tracking or dialogue state tracking model could be used to maintain the dialogue flow \cite{ultes2017pydial}.
We used a rule-based model where the dialogue flow module keeps track of the input dialogue acts and DoP and send them to the response manager to fetch responses.
The complete state loop has a queue to store the context information of the preceding utterances.
It is helpful to trigger new dialogue acts based on the context information. 
For example, if the last dialogue act is \textit{TakeToPlace}, it triggers a new dialogue act called \textit{FinishedOne} to inform the system that the last action was finished and asks the user if he/she wishes to visit the next place.
Another loop keeps track of whether the user accepts or rejects the proposal using \textit{Accept} and \textit{Reject} dialogue acts.
If one of the dialogue acts appears, the robot takes the user to the next location until either the list of locations is finished or the user rejects to visit the next place.

\subsection{Response Manager}

The response manager is responsible for picking up the right response for the given intention and degree of politeness.
Pre-defined response templates are stored in a data file that is accessed continuously during the interaction.

\section{Robot Control and Navigation}

\subsection{Robot Platform}
Pepper is a 1.2 meter tall omnidirectional wheeled humanoid robot platform capable of exhibiting body language, perceiving and interacting with its surroundings, and move autonomously. 
Due to its 17 joints and 20 degrees of freedom (DoF) kinematic configuration and edgeless design, the system is suitable for safe HRI \cite{pandey2018pepper}.
The platform is equipped with a large variety of sensors and actuators that ensure safe navigation and a high degree of expressiveness: LED's are distributed across the head (eyes and ears) and torso (shoulders) to support non-verbal communication by modifying colour and intensity. 
The microphones and speakers allow verbal interaction as well as environmental awareness. 
Sensing components include three laser sensors, two sonars and two infrared sensors located in the robot's base, as well as two cameras and a three-dimensional camera located in the head. 
%In addition, two tactile detectors on the back of both hands allows human-robot physical awareness. 
Finally, the platform is powered by an Atom processor with a 1.91 GHz quad-core unit that allows the NAOqi SDK to orchestrate the different hardware elements as well as their access from other APIs.

\begin{figure}[t]
\includegraphics[width=9.5cm]{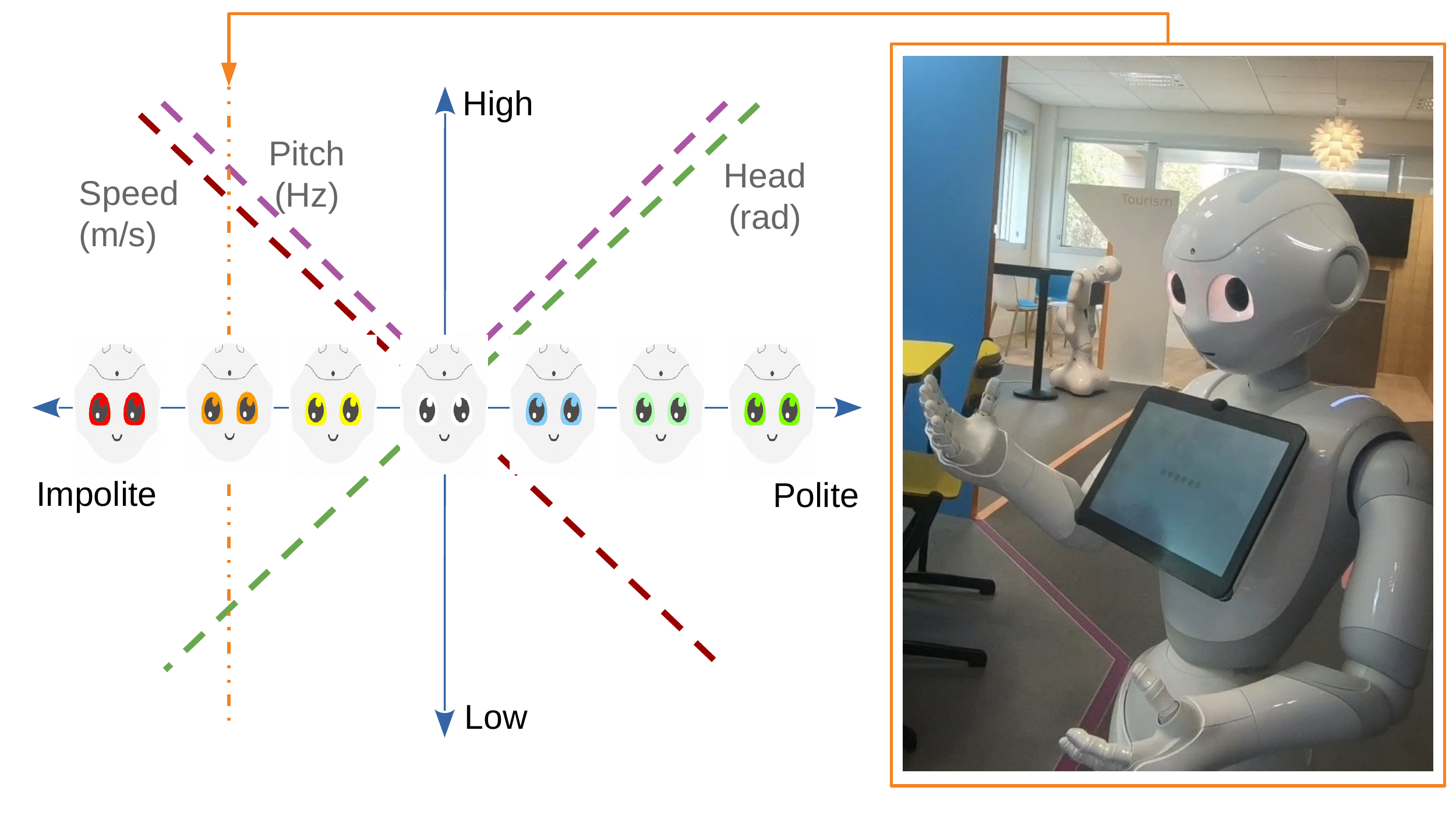}
\centering
\caption{The behavioural model used to create the verbal and non-verbal responses based on the cumulative sum of the DoP. 
%Head pitch position raises as DoP increases whereas the robot's navigation speed does the opposite. 
%In the case of voice pitch, responses increase equally for both extremes. 
%Eyes color code matches the robot's internal emotional state. 
The Pepper robot shown in the right is in the position of the vertical orange line in the plot during the interaction.} 
\label{fig:state_manager}
\end{figure}

\subsection{State Manager}

In order to produce the physical and verbal responses in accordance with the degree of politeness exhibited during the interaction, a behavioural model inspired by the valence and arousal model \cite{beck2010towards} has been designed. 
The model is given the discrete DoP computed from the last utterance being [1, 0, -1] and maps the cumulative sum of the politeness of the previous and current utterances to different actuators. 
In this way, a variability is provided to every single social cue that can vary in order to fit the interaction needs.

The actuators used to externalize the robot's change of state are the LED's color \cite{nijdam2009mapping}, head pitch orientation \cite{lemaignan2016real}, voice pitch \cite{hubbard2017production} and navigation speed, and are mapped following the intuition (shown in Figure \ref{fig:state_manager}). 
For example, a user repetitively polite during the whole interaction will experience a decrement in the navigation speed of the robot, a head position oriented towards the user, green coloured eyes and a slightly higher voice pitch.

\subsection{Motion Manager}
The motion manager is responsible for navigation and can be operated in the following modes:
%The motion manager is composed of the following three modules.
%: Tele-operation, Scripted Navigation and Navigation with Mapping.
%scripted is used for testing
%TODO: Arturo -- input about motion manager: with the future projection about Navigation planner, also add subsubsection if needed
\vspace{-10pt}
\subsubsection{Tele-operation}
In this mode, the Pepper robot could be teleoperated with the help of the NAOqi framework using the \textit{moveToward} function from ALMotion service and the keys on the keyboard are used for moving or stopping the robot. 
\vspace{-20pt}

\subsubsection{Scripted Navigation}
The scripted navigation is achieved by commanding a robot to move to the specific positions/places with the known distances in the environment. 
This is also achieved with NAOqi framework using \textit{moveTo} command from ALMotion service.
We specify how far the robot has to move (in meters) and the orientations (in radians) it has to take during motion.
\vspace{-10pt}

\begin{figure}[t!]
\includegraphics[width=9cm]{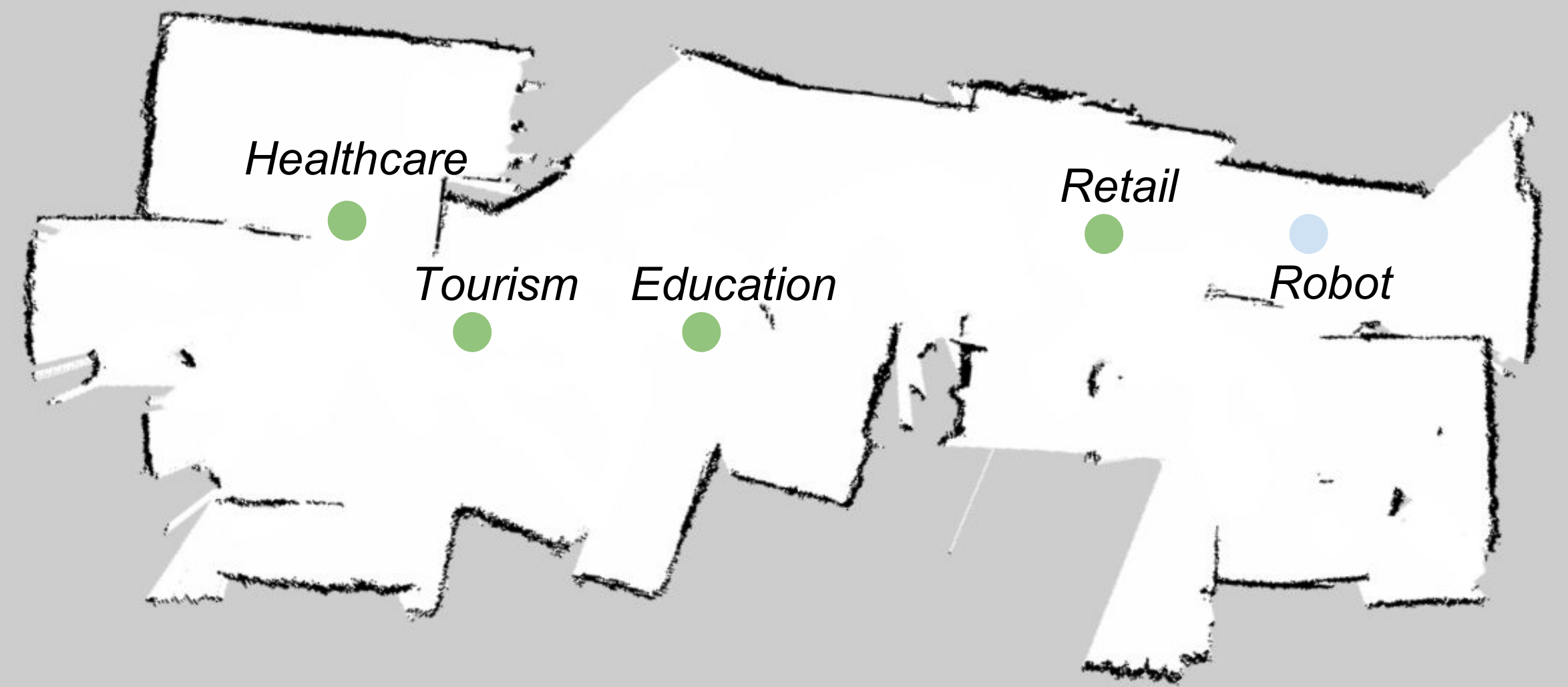}
\centering
\caption{The environment map created with the Pepper robot and gmapping from ROS. }  % depth sensor
\label{fig:map}
\vspace{-10pt}
\end{figure}

\subsubsection{Navigation: Mapping and Planning}
This module requires the use of the Robot Operating System (ROS), an open source middle-ware framework. 
To fit our need for navigation, we have adopted the following approach for generating and post-processing the map.
The current readings of the Pepper's depth image are converted into virtual laser data, using the package \textit{depthimage\_to\_laserscan} \cite{perera2017setting,suddrey2018enabling}. 
An offline map (shown in Figure \ref{fig:map}, and post-processed for testing purposes) can be acquired using \textit{gmapping} (laser-based SLAM) \cite{grisettiyz2005improving}. 
%The map was generated using the Pepper robot with only the virtual laser data. % without lenses on the eyes.
Then, the localization is performed using Adaptive Monte Carlo Localization (\textit{acml}) \cite{fox2003adapting}. 
Finally, the navigation uses a global planner with a map with inflated obstacles (costmap) and a local costmap with observations from the virtual laser data. 
The Dialog Flow requests a location from the API server (on the robot using a virtual machine) using an ID and this one sends the coordinates to the ROS navigation stack to execute the path. 

\vspace{-10pt}
\section{Experiments and Results: A Real-World Scenario}

\begin{figure}[t!]
\includegraphics[width=9cm]{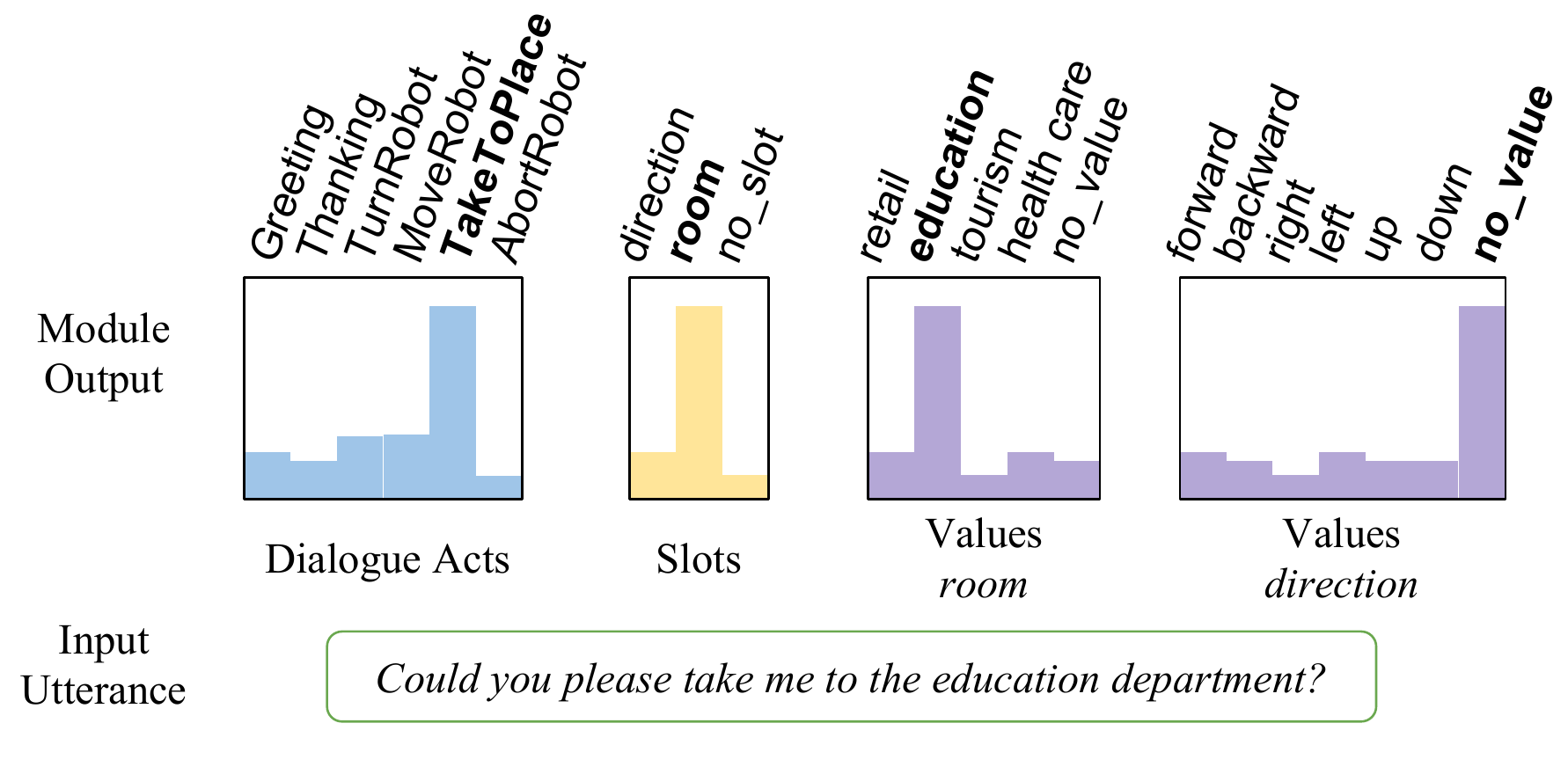}
\centering
\caption{Output of the DA recognition module. } 
\label{fig:slu_example}
\end{figure}

The task is to navigate in the given environment to show a user the different departments.
Our tour scenario in the lab consists of four departments: retail, education, tourism and healthcare as shown in the map in Figure \ref{fig:map}.
The robot is a guide which takes the user to the particular department using verbal interaction as mentioned in Section \ref{section:dialogue_flow}.
%For the testing purpose, we used the scripted navigation module.
When the user asks the robot, the input utterance gets processed by the DA recognition module which produces the result as shown in Figure \ref{fig:slu_example}.
The politeness detection module provides the DoP of that utterance.
The dialogue flow communicates this information with all the managers. 
The robot adapts its behaviour such as speeding up or down while navigating to the locations and changing the pitch of speech, changing the pitch angle of the head.

\begin{figure}[b!]
  \centering
  \subfigure[]{\includegraphics[width=0.45\textwidth]{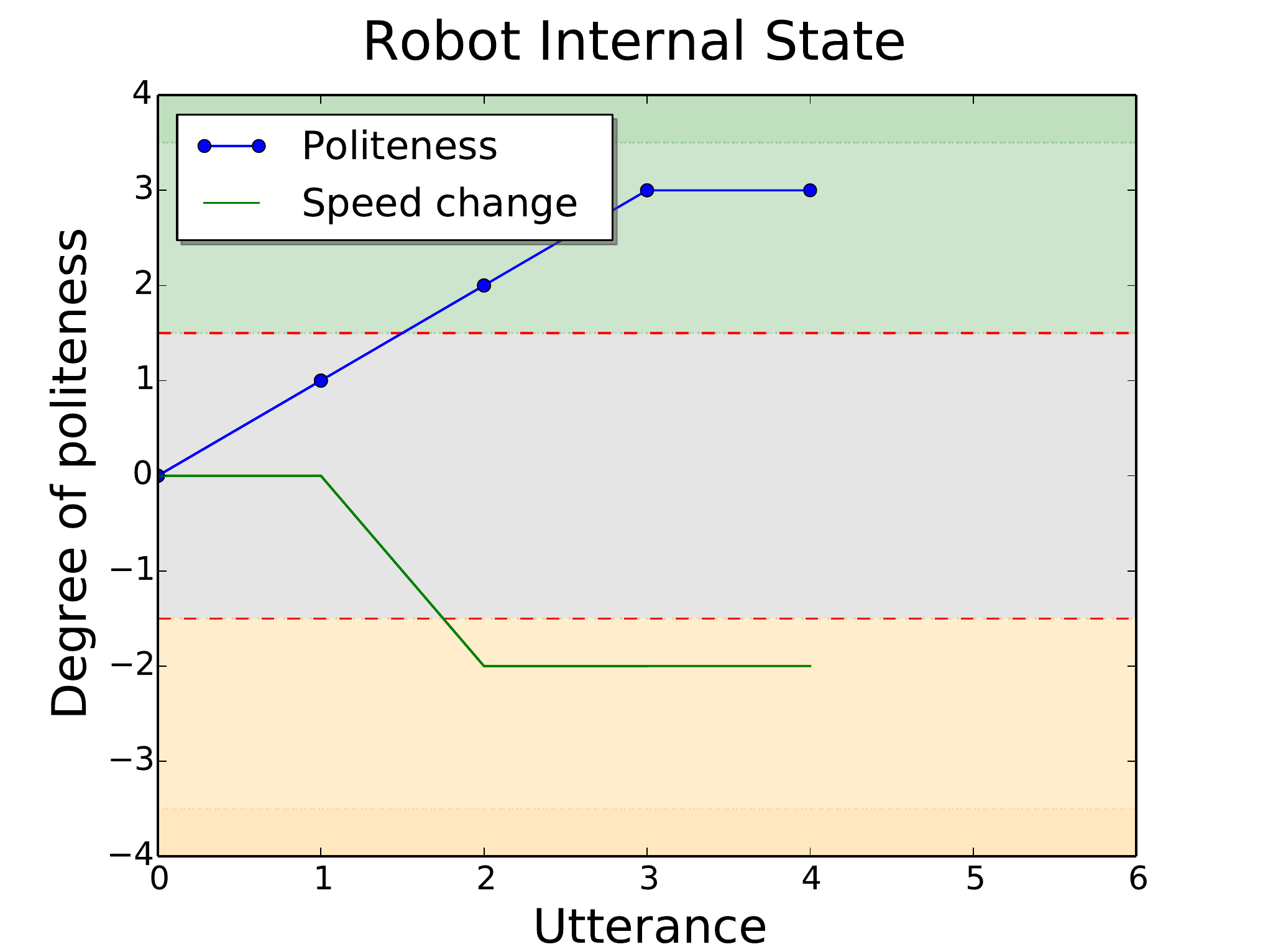}}
  \subfigure[]{\includegraphics[width=0.45\textwidth]{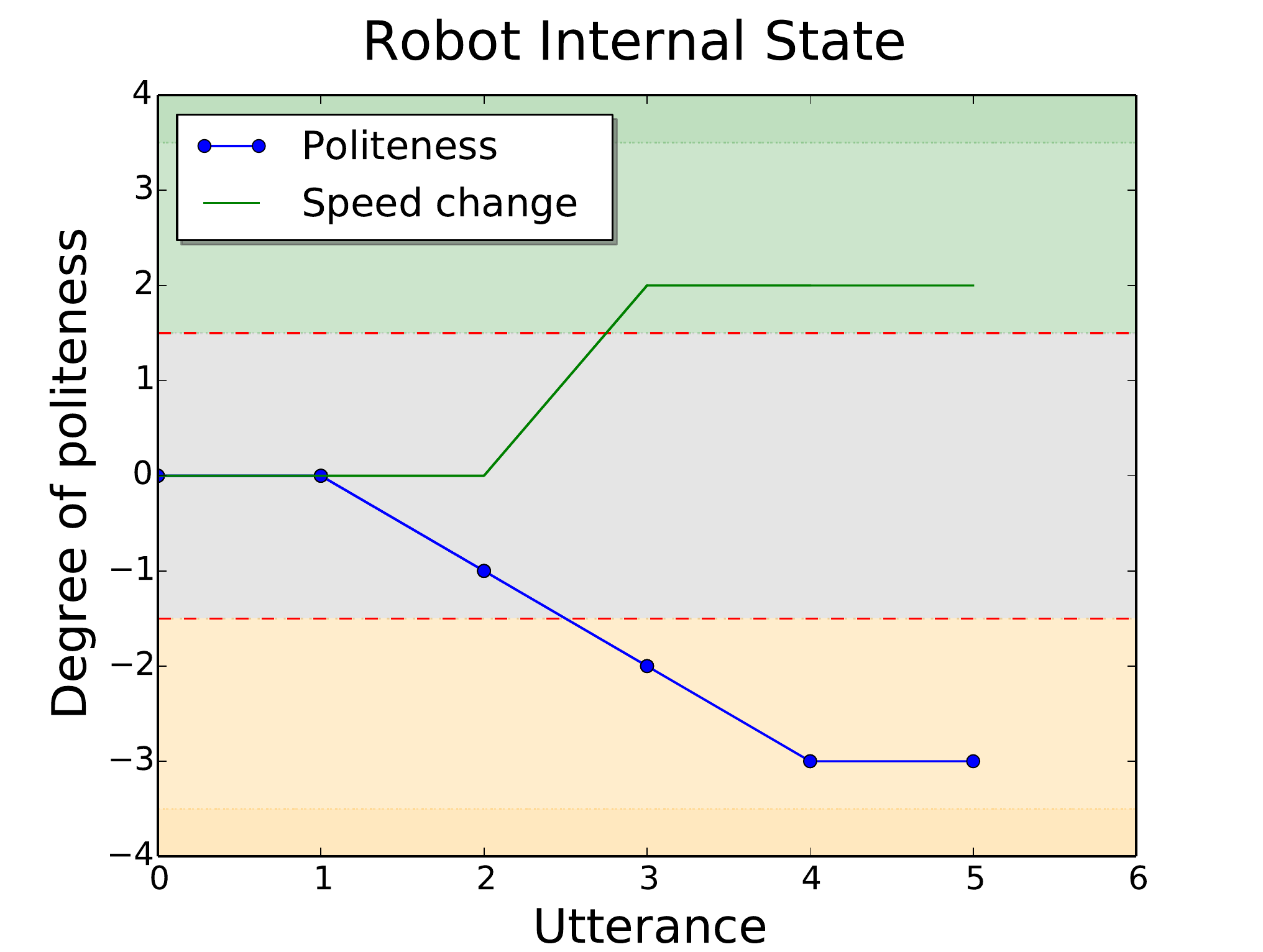}}
  \caption{Robot internal state for polite (a) and impolite (b) interactions.}
  \label{fig:polite_impolite}
  \vspace{-30pt}
\end{figure}

We tested our system on the Pepper robot with different users, expressing different levels of politeness. 
The behavioural changes and adaptation to speed change based on a change in DoP are shown in Figure \ref{fig:polite_impolite}.
The robot behavioural adapts to the human being polite; the robot slows down and spends more time with the user. 
When the user is impolite, the robot speeds up and executes motion faster.
The proposed behaviour of the robot for different situations shown in the figure is mainly to demonstrate the developed system and the efficacy of the proposed framework. 
The results indicate that the system is able to consider the linguistic features to modulate the navigation behaviour of the robot in a coherent theoretical and functional framework. 
As aforementioned, to the best of our knowledge such a framework and implementation in a practical situation is one of the first attempts of its kind.
However, it is important to mention that the validation of the hypotheses about the most appropriate behaviours of the robot is not within the scope of this paper and it will require further investigation and user studies. 
As mentioned in the conclusion such studies are one of the next steps to utilize the framework for different situations.
The demonstration video and dialogue logs of the generated graphs in Figure \ref{fig:polite_impolite} are available at the SECURE EU Project website: \url{https://secure-robots.eu/fellows/bothe/secondment-project/}

%\subsubsection{Polite Scenario}
%\subsubsection{Impolite Scenario}
\vspace{-10pt}

\section{Conclusions and Future Work}

We developed a dialogue-based navigation system for integrating intention and politeness features for multivariate adaptation of the robot. 
We successfully deployed and tested our system on the robot with different levels of politeness. 
Currently, our work does not elicit the causal explanation for the behaviour and the multivariate adaptation of the robot.
However, our experimental framework opens up a new challenge for the study of the effect of politeness in human-robot social interaction.
We strongly believe that our work will be helpful in bridging the gap between sociolinguistic research and the HRI community. 
This research shall also be helpful in targeting the deployment of social-service robots with adaptation to sociolinguistic features such as politeness.
In this work, the behaviours are based on previous research \cite{fong2003survey,manav2007color,salem2014marhaba,srinivasan2016help}. %, however, 
%the user study would suffice the demonstration of the effect of politeness.
%Our main focus was on language processing, but, other modalities could be added such as gesture recognition, facial emotion recognition, etc.
The validation of the system is crucial and it will be addressed in future work through user studies.

% If YOU some bibtex to add, Please add them to lokalbib.bib file
%
\vspace{-10pt}

\section*{Acknowledgements}
\vspace{-10pt}

This project has received funding from the European Union's Horizon 2020 framework programme for research and innovation under the Marie Sklodowska-Curie Grant Agreement No. 642667 (SECURE), the Industrial Leadership Agreement (ICT) No. 779942 (CROWDBOT), and No. 688147 (MuMMER).
\vspace{-10pt}

\bibliographystyle{splncs04}

%\bibliography{mybib,emnlp2018,lokalbib}

\end{document}